\documentclass[journal]{IEEEtran}
\usepackage{graphicx}
\usepackage{float}
\usepackage{subcaption}
\usepackage{amsmath}
\usepackage{bm}
\usepackage{upgreek}
\usepackage{color}
\usepackage{gensymb}
\usepackage{cite}
\usepackage[numbers,sort&compress]{natbib}
\usepackage[left=0.75in,right=0.75in,top=1in,bottom=0.8in]{geometry}
\usepackage{diagbox}
\usepackage{algorithm}  
\usepackage{algpseudocode}  
\usepackage{amsmath}  
\ifCLASSINFOpdf
\else
\fi
\begin{document}

\title{Current Effect-eliminated Optimal Target Assignment and Motion Planning\\ for a Multi-UUV System}
%
%
% author names and IEEE memberships
% note positions of commas and nonbreaking spaces ( ~ ) LaTeX will not break
% a structure at a ~ so this keeps an author's name from being broken across
% two lines.
% use \thanks{} to gain access to the first footnote area
% a separate \thanks must be used for each paragraph as LaTeX2e's \thanks
% was not built to handle multiple paragraphs
%

\author{Danjie~Zhu~and~Simon~X.~Yang% <-this % stops a space
\thanks{The authors are with Advanced Robotics and Intelligent System (ARIS) Laboratory, School of Engineering, University of Guelph, Guelph, ON. N1G2W1, Canada (e-mail: danjie@dal.ca, syang@uoguelph.ca).}
}% <-this % stops a space
\maketitle

% As a general rule, do not put math, special symbols or citations
% in the abstract or keywords.
\begin{abstract}
The paper presents an innovative approach (CBNNTAP) that addresses the complexities and challenges introduced by ocean currents when optimizing target assignment and motion planning for a multi-unmanned underwater vehicle (UUV) system. The core of the proposed algorithm involves the integration of several key components. Firstly, it incorporates a bio-inspired neural network-based (BINN) approach which predicts the most efficient paths for individual UUVs while simultaneously ensuring collision avoidance among the vehicles. Secondly, an efficient target assignment component is integrated by considering the path distances determined by the BINN algorithm. In addition, a critical innovation within the CBNNTAP algorithm is its capacity to address the disruptive effects of ocean currents, where an adjustment component is seamlessly integrated to counteract the deviations caused by these currents, which enhances the accuracy of both motion planning and target assignment for the UUVs. The effectiveness of the CBNNTAP algorithm is demonstrated through comprehensive simulation results and the outcomes underscore the superiority of the developed algorithm in nullifying the effects of static and dynamic ocean currents in 2D and 3D scenarios.
\end{abstract}

% Note that keywords are not normally used for peerreview papers.

% For peer review papers, you can put extra information on the cover
% page as needed:
% \ifCLASSOPTIONpeerreview
% \begin{center} \bfseries EDICS Category: 3-BBND \end{center}
% \fi
%
% For peerreview papers, this IEEEtran command inserts a page break and
% creates the second title. It will be ignored for other modes.
\IEEEpeerreviewmaketitle

\section{Introduction}
%%%%%%%%%%%%%%%%%%%%%%
\IEEEPARstart
{T}{he} studies on unmanned underwater vehicles (UUV) have been attractive for many researchers in recent decades due to the flexibility of the vehicle in the underwater area \cite{r5,r6}. To overcome the difficulty of accomplishing complex underwater operations, the multi-UUV system, which refers to the system of multiple UUVs and multiple targets, has attracted great attention owing to its high parallelism, robustness and efficiency \cite{petillo2014,Panda2020,Hadi2021}. In a multi-UUV system, the target assignment and motion planning problem is the most basic and significant issue as it decides the efficiency of the UUV collaboration \cite{Cai2017}. The most optimal target assignment and motion planning in the multi-UUV system is realized by deducting the solution that consumes the shortest total distance or the shortest time.

Algorithms regarding the optimal target assignment and motion planning problem for UUVs have been investigated by many researchers, where most algorithms focus on path planning and rely on the grid-based modeling of the environment \cite{3dpp,ppcon,pptime,uavcurrent2}. Dijkstra algorithm is one of the earliest grid-based path planning methods where a global search on all possible path solutions is required \cite{Dijkstra}. A* algorithm is then raised with the advancement of adding the heuristic cost to reduce the searching space \cite{Astar1}. However, when considering the disturbances in various kinds of environment, most studies focus on the wind effect of the unmanned aerial vehicle (UAV), instead of the underwater environmental factors \cite{Lin2017,wind4,wind2,wind1}. The typical underwater disturbances, for example, the effect of currents and obstacles usually have an inevitable influence on the UUV operation process due to their irregular distribution and low visibility of the marine environment \cite{RN1345}. Therefore the traditional grid-based path planning methods that need maps of high accuracy and consistency are not appropriate to the UUV system \cite{RN1338}.  

The A* algorithm has been adapted for application in the marine system, incorporating environmental influences such as moving obstacles and ocean currents \cite{RN1337, RN1338}. While feasible methods have been proposed for maintaining safety and efficiency during optimal path tracking, these approaches are primarily limited to unmanned surface vehicles (USVs), which require fewer dimensions. However, the use of the A* algorithm on UUVs is challenging due to their limited velocity and operating space, necessitating real-time feedback for navigating complex undersea environments \cite{rr3}. To address this issue, the genetic algorithm (GA) has been combined with path planning for UUVs, as it allows for regulation of optimization based on environmental factors \cite{EA1, RN1346, ant2}. Nonetheless, the high computational complexity and large data inputs associated with the GA method limit its applicability in UUV path planning \cite{EA2}. This challenge is particularly evident in multi-UUV systems, where efficient target assignment plays a crucial role in managing multiple tasks simultaneously and improving system efficiency \cite{Wu2021}.

Various intelligent algorithms have been developed to address the efficient assignment of targets and planning for multiple UUVs with multiple targets, with the self-organizing map (SOM) algorithm being widely accepted \cite{Huang2013, Cao2018, Mingzhi2019, Wu2021, Hung2021}. The SOM algorithm has been employed in multi-UUV systems due to the similarity between the target assignment and SOM network assignment. However, the collaboration efficiency among UUVs still has room for improvement, and the impact of ocean currents is often not considered \cite{bioins, Ma2021}. Moreover, studies that combine deep learning (DL) and reinforcement learning (RL) have been applied in ground vehicle planning problems, where a recurrent deep neural network is designed to perform an efficient convergence of the optimized planning results \cite{Chai2023,Chai2020,Chai2021,Chai2022,ChaiTASE2023,ChaiTIE2021}. Nevertheless, in underwater conditions, the high computation required by the recurrent neural network cannot be perfectly realized due to communication limitations and in-time feedback requests. Therefore, there still exists room for improvement in collaboration efficiency and the consideration of environmental factors such as ocean currents. Future research could focus on developing algorithms that address these limitations and enhance the overall performance of UUVs in complex undersea environments \cite{Bayat2016,Mengjia2022}.

Motivated by the challenges of eliminating the effect of ocean currents and achieving efficient collaboration in multi-UUV systems, the bio-inspired neural network (BINN) is proposed as a solution for target assignment and motion planning in the marine environment. Unlike other intelligent methods such as A*, GA, SOM, or DL/RL-based methods, BINN provides real-time feedback on underwater environmental factors, including current effects and obstacles, without imposing strict requirements on input data. This is achieved by continuously updating the state of neurons and transmitting information throughout the network, enabling instant reactions and reducing computational complexity by limiting the search area within a certain range \cite{bioins, RN1348}. The acceptable computational complexity and the real-time reaction of the BINN-based approach make it an appropriate choice for applications in multi-UUV systems. Additionally, the application of BINN facilitates the decomposition of current effects at each neuron, allowing for the efficient elimination of current influences. Moreover, the BINN method enhances target assignment efficiency by simultaneously generating target attraction while addressing optimal paths.

This paper presents an intelligent algorithm based on a BINN for target assignment and motion planning in multi-UUV systems operating in a marine environment with the current effect involved. The second section of the paper provides a detailed explanation of the proposed method. Firstly, BINN is utilized to establish an underwater map model and determine the optimal path that ensures the shortest distance and avoids collisions. Additionally, an algorithm is introduced to compensate for the deviation caused by ocean currents, thus enabling the UUVs to navigate along the optimal path. Furthermore, the paper describes the target assignment process for multiple UUVs based on the path distances obtained from the proposed algorithm, where the distance metric reflects the priority of task efficiency in practical scenarios. In the third section, the paper presents comprehensive simulation results. It includes a detailed analysis and comparison of the algorithms, considering both the scenarios where the existence of ocean currents is considered and those where it is not. The simulation results highlight the effectiveness of the proposed algorithm in terms of target assignment and motion planning, especially in current-affected marine environments.

%%%%%%%%%%%%%%%%%%%
\section{Problem Formulation}
In this section, the model of the underwater motion planning and target assignment problem for a multi-UUV is established and explained. 
\subsection{Map Modeling of the Underwater Enviornment}
The underwater environment under 2D condition of the multi-UUV system is modeled as a two-dimensional grid-based map, where the grids are decomposed as small enough to represent the irregular obstacles through two conditions: fully occupied (in dark) or not (in white), as shown in Fig. 1. Additionally, the exact position of the vehicle can be addressed by the 2D coordinates established on the grid map, where the side length of each grid covers a single unit of the system. The red grids represent the target positions while the blue grids represent the vehicle positions, with their exact coordinates assured in the modeling map.

\begin{figure}[t]
\begin{center}
    %\begin{flushmiddle}
        \includegraphics[scale=0.37]{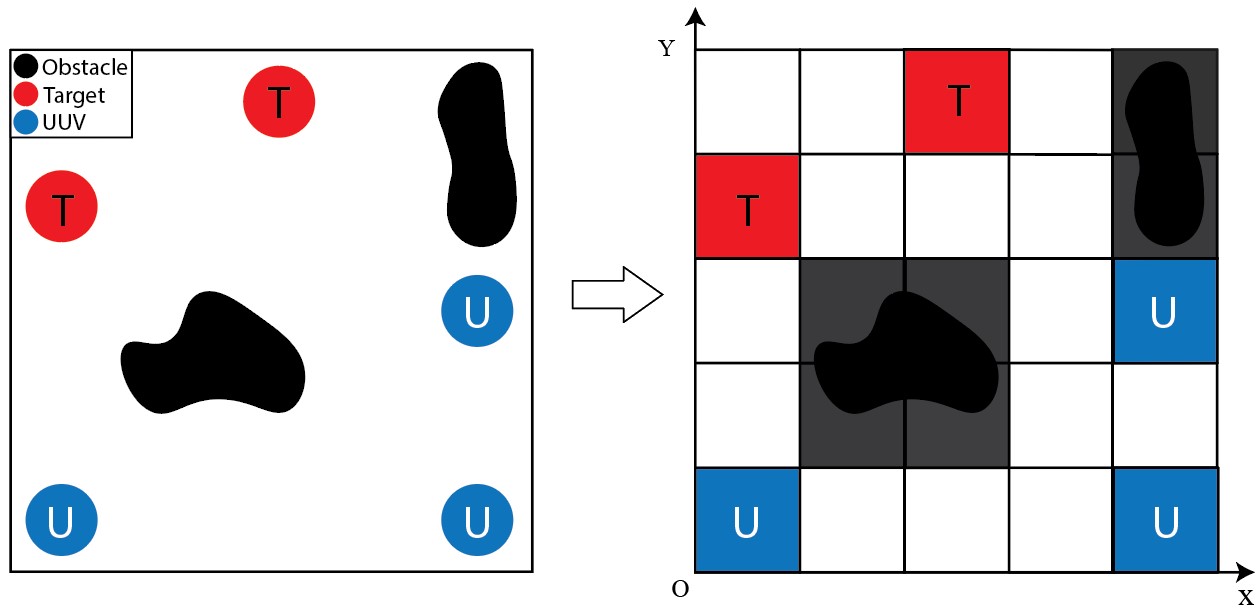}                         %\captionsetup{justification=centering}
        \caption*{Fig. 1. 2D grid-based map modeling for the system of multiple UUVs and targets in the underwater environment}	
    %\end{flushmiddle}
\end{center} 
\end{figure}  %%%%%%true figure

\subsection{Goals and Constraints}
The problem of target assignment and motion planning in a multi-UUV system involves determining the optimal allocation of targets among a group of autonomous agents and finding the most efficient paths for each agent to reach their assigned targets, which usually in a way of maximizing the overall system performance, such as shortest total time or distance consumption.

The targets can represent various tasks, locations, or objectives that need to be accomplished by the UUVs. The primary objectives of the target assignment include maximizing the number of completed tasks, minimizing the overall completion time, balancing the workload among UUVs, or optimizing some other performance metric. Motion planning is another crucial aspect of the problem, where each UUV needs to determine the most efficient route to reach its assigned target while avoiding obstacles, collisions with other agents, and considering any constraints or limitations such as speed, fuel consumption, or energy resources. In this study, we do not consider the collisions with other agents while the limitations of the UUV such as speeds are involved.

The complexity of the target assignment and motion planning problem increases with the number of agents, targets, and the dynamic nature of the environment. The challenges include handling uncertainty, dealing with incomplete or changing information, accounting for inter-agent communication and coordination. Therefore, in this paper, we apply a multi-UUV system of less than 5 agents to perform a more direct result showing the effectiveness of the proposed target assignment and motion planning method. 

The influence of ocean currents also poses a significant challenge for multi-UUV motion planning. Ocean currents have the potential to disrupt the intended course and trajectory of UUVs, bringing difficulties in maintaining precision in positioning or reaching specific waypoints. This study addresses both static and dynamic forms of ocean currents, each presenting distinct challenges. Static currents characterized by linear changes in direction and flow rates are considered, which can be factored into UUV motion planning, requiring adjustments to account for their predictable effects on vehicle paths. Moreover, dynamic currents introduce nonlinear variations, leading to complex and unpredictable flow patterns like eddies, vortices, and turbulent regions. These dynamic currents can induce unexpected deviations from planned courses, making the planning and execution of UUV navigations more demanding.

In summary, the goal of the study is to address efficient target assignment and motion planning results for a multi-UUV system with the shortest possible distances in total. Corresponding constraints are applied, such as obstacles and ocean currents in both static and dynamic forms.
%\hfill mds
 
%\hfill August 26, 2015
%%%%%%%%%%%%%%%%%%%%%%

%%%%%%%%%%%%%%%%%%%%%%
\section{Current Effect-eliminated Bio-inspire Neural Network-based Target Assignment and Motion Planning (CBNNTAP) Algorithm Design}
In this section, the proposed algorithm based on a BINN is described to resolve the motion planning and target assignment problem for a multi-UUV system, which is divided into three parts: the BINN-based path planning component; the target assignment strategy derived from the path planning results; and the current effect elimination component that realized during the navigation according to the BINN model \cite{DJZthesis}.
\subsubsection{Bio-inspired Neural Network Path Planning (BNNP) Component}
A neural network is built according to the grid-based map, where each grid of the map represents a neuron in the network respectively, for example, the $i^{th}$ grid represents the $i^{th}$ neuron (Fig. 2). The information carried by each neuron can be transmitted to the eight adjacent neurons within the receptive range that is presented as a circle of $\sqrt{2}$ radius in Fig. 2. Further speaking, once a neuron is activated, the eight adjacent neurons are going to be activated and accept the transmitting information. One of the eight neurons will then be activated based on the algorithm and repeat the next activation loop on the eight adjacent neurons. The activation and transmission process exactly fulfills the requirement of the UUV path planning on a grid-modeling map. Once the vehicle reaches a grid, one of the eight adjacent grids will be chosen as the next position along the path. The choice of the next activated neuron as well as the next position grid is addressed by the intelligent algorithm proposed in this paper.

Following a similar definition, in the 3D neural network, each neuron is represented by a cube, where the information carried by each neuron can be passed to the 26 adjacent neurons that belong to the 26 adjacent cubes according to the 3D structure. A sphere whose radius is $\sqrt{2}$ that centers at each neuron covering the 26 adjacent neurons defines the receptive range of the 3D network. One of the 26 neighboring cubes will be the next location for the vehicle.

\begin{figure}[t]
\begin{center}
    %\begin{flushmiddle}
        \includegraphics[scale=0.235]{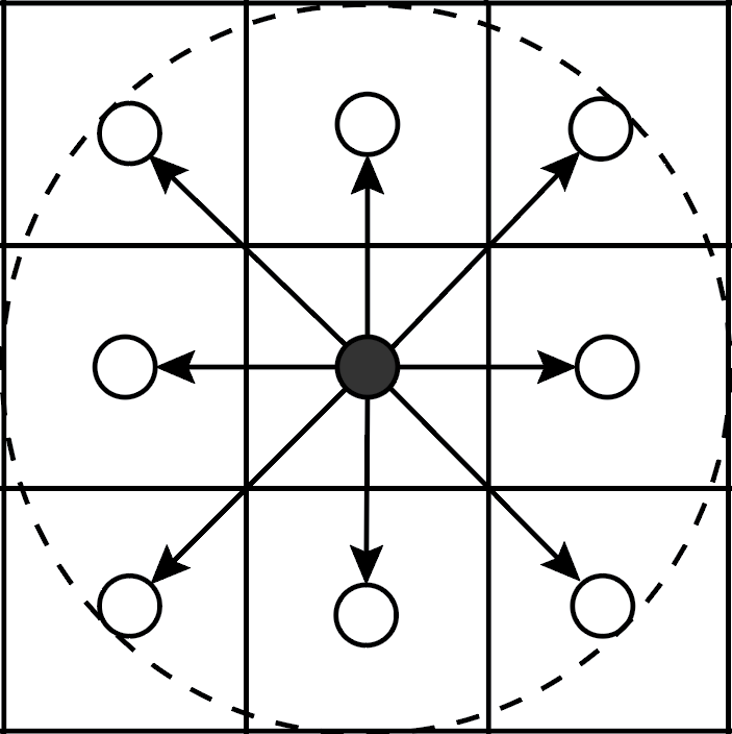}                         %\captionsetup{justification=centering}
        \caption*{Fig. 2. The 2D bio-inspired neural network}	
    %\end{flushmiddle}
\end{center} 
\end{figure}  %%%%%%true figure
Once the grid-based neural network model is established, the CBNNTAP method for motion planning and target assignment of a multi-UUV system can be proposed. The CBNNTAP algorithm is mainly formed by three components (Fig. 3): (1) a BNNP algorithm component that deducts the optimal path by avoiding all the obstacles; (2) a target assignment component based on the path distances given by the BNNP algorithm; (3) an elimination component that balances off the deviation induced by the effect of ocean currents.
\begin{figure*}[t]
\begin{center}
    %\begin{flushmiddle}
        \includegraphics[scale=0.45]{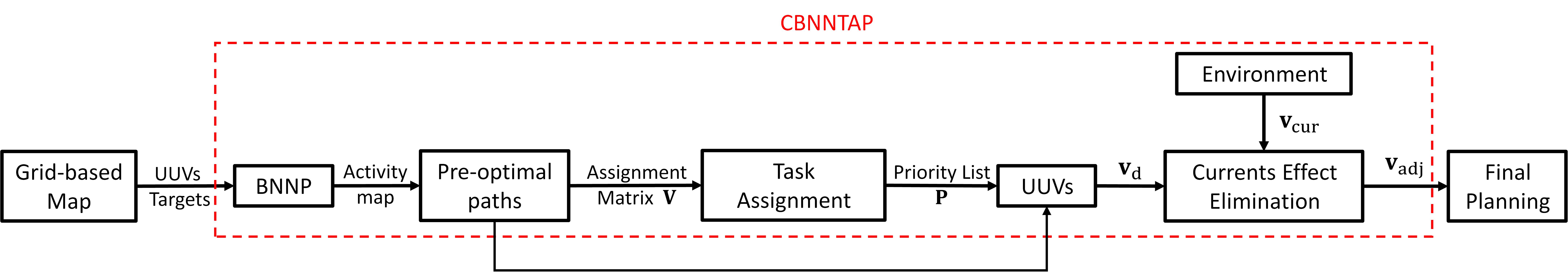}                         %\captionsetup{justification=centering}
        \caption*{Fig. 3. Schematic of the CBNNTAP method designed for a multi-UUV system}	
    %\end{flushmiddle}
\end{center} 
\end{figure*}   %%%%%%true figure
%%%%%%%%%%%%%%%%%%%%%%

%%%%%%%%%%%%%

The BNNP algorithm is developed on the grid-based map and the corresponding neural network model, based on a discrete-time Hopfield-type neural network algorithm \cite{Zhu2021}. The algorithm is to derive the most optimal path with the shortest distance and no collisions, which is formed by the continuous coordinates of the vehicle movement. The start position and target position are known and marked on the map at the beginning, where the position and shape of the obstacles are unknown until detected by the vehicle during the planning process. Therefore, the dynamic model of the BNNP algorithm which deducts the next position of the planning path can be defined as
\begin{equation}
    a_i(t+1)=f(a_j(t)+e^{(-||i-T||+||i-j||)}+J_i),
\end{equation}
where $a_i (t+1)$ represents the neural activity of the $i^{th}$ neuron at the time $t+1$; $a_j (t)$ represents the neural activity of the $j^{th}$ neuron at the time $t$.

At each position, the neighboring neuron with the highest activity will be chosen. Under the condition of two dimensions, the number $i$ ranges from 1 to 8, referring to the eight adjacent neurons of the activated $j^{th}$ neuron within the predefined receptive range (see Fig. 2). Under the condition of three dimensions, the number $i$ ranges from 1 to 26, representing the twenty-six adjacent neurons around the activated $j^{th}$ neuron within the receptive scope of a sphere whose radius is $\sqrt{2}$ that centers at $j^{th}$ neuron. Additionally, $||i-T||$ is the Euclidean distance between the $i^{th}$ neuron and the Target T. The part of $e^{(-||i-T||+||i-j||)}$ represents the closer the vehicle approaches the target at the next position, the higher activity the corresponding neuron will have.

The transfer function $f(x)$ is
\begin{equation}
    f(x)=\left\{
    \begin{aligned}
    -1, \quad x<0\\
    k_fx, \quad x\geq0
    \end{aligned}
    \right.,
\end{equation}
where $k_f$ is a constant chosen between 0 and 1. In this paper we assign $k_f$ as 0.5.

The external input to the $i^{th}$ neuron, $J_i$, is determined according to the condition of the obstacle occupancy (fully occupied or not) and the completeness of the searched status for the current neuron as
\begin{equation}
    J_i=\left\{
    \begin{aligned}
    -1, \quad \text{if obstacles}\\
    0, \quad \text{if searched}\\
    +1, \quad \text{if unsearched}
    \end{aligned}
    \right.,
\end{equation}
where $J_i$ represents the external input to the $i^{th}$ neuron.

On the 2D map, at the time when the vehicle is going to choose its next position, the eight adjacent units around it will be the next possible transferring neurons, where their coordinates can be obtained by adding or subtracting 1 from the vehicle’s current coordinate. On the 3D map, one of the twenty-six units around the currently activated neuron will be the next possible local solution following a similar rule. The network's activated neurons are continuously updated as the vehicle moves, resulting in the development of an activity map, which guides the determination of an optimal path solution.

%%%%%%%%%%%%%%%%%%%%%%
\subsubsection{Target Assignment Component}
The target assignment in the system of multiple vehicles and targets is addressed by the path distances of the pre-optimal path results deducted by the BNNP algorithm. For the pre-optimal path planning process, the multiple targets are separately assigned to each UUV, with the optimal path of each possible task between a UUV and a target derived by the BNNP algorithm. For example, the pre-optimal path planning results in a $10\times10$ map with two targets and three UUVs are shown in Fig. 4, where all six possible paths are computed.
\begin{figure}[!b]
\begin{center}
    %\begin{flushmiddle}
        \includegraphics[scale=0.34]{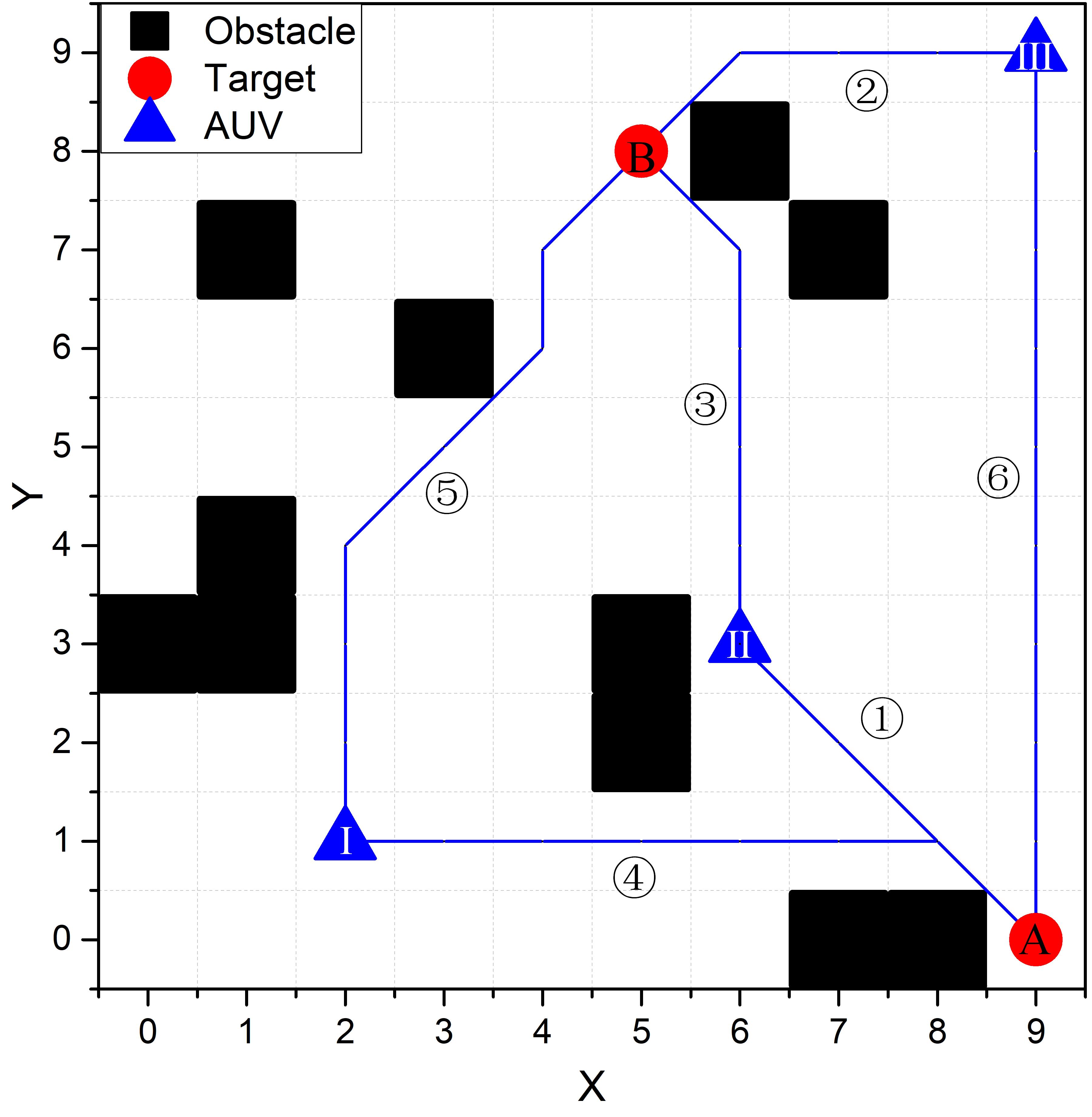}                         %\captionsetup{justification=centering}
        \caption*{Fig. 4. Pre-optimal path planning for a multi-UUV system}
    %\end{flushmiddle}
\end{center} 
\end{figure}  %%%%%%true figure

The priority of accomplishing the tasks is assessed by the normalized path distances and the ratio of each path distance and the longest path distance is recorded as an Assignment matrix $\mathbf{A}$ such that the matrix value can be limited between 0 and 1 to show a clear comparison. The assignment matrix $\mathbf{A}$ is defined as
\begin{equation}
    \mathbf{A}=
    \begin{bmatrix}
    A_{11} & A_{12} &...& A_{1n}\\
    A_{21} & A_{22} &...& A_{2n}\\
    . & .& & .\\
    . & & .& .\\
    . & & & .\\
    A_{m1} & A_{m2} &...& A_{mn}\\
    \end{bmatrix},
\end{equation}
where $m$ means the number of targets, $n$ represents the number of vehicles.

Once the assignment matrix $\mathbf{A}$ is addressed, the path of the lowest value in the matrix is first chosen as the prior task and the corresponding UUV is considered as the winner to the target of the chosen path. The row and column of the chosen value are then eliminated to form a new matrix, which is noted as the matrix residue. The choice of the winner UUV to the target follows the rule of continuously picking the lowest value in the residue matrix and kicking the corresponding row and column to achieve the new residue matrix until all the targets are assigned. The Priority list is defined as,
\begin{eqnarray}
&&\mathbf{P}=[P_1, P_2, ..., P_m] \notag\\
&&=[min(\mathbf{A}), min(\mathbf{A_{p1}}), ..., min(\mathbf{A_{pm-1}})],
\end{eqnarray}
where $\mathbf{P}$ represents the priority list of target assignment that contains the ordered values chosen from $\mathbf{A}$; $A_{p1}$ and $A_{pm-1}$ represents the residue matrix derived from the former assignment step. Supposing $A_{ij}=min(\mathbf{A})$, the residue matrix can be defined as
\begin{equation}
    \mathbf{A_{p1}}=
    \begin{bmatrix}
    A_{11} &...& A_{1(j-1)} & A_{1(j+1)} & A_{1n}\\
    A_{21} &...& A_{1(j-1)} & A_{1(j+1)} & A_{2n}\\
    . & .\\
    . & & .\\
    . & & & .\\
    A_{(i-1)1} &...& A_{1(j-1)} & A_{1(j+1)} & A_{(i-1)n}\\
    A_{(i+1)1} &...& A_{1(j-1)} & A_{1(j+1)} & A_{(i+1)n}\\
    . & .\\
    . & & .\\
    . & & & .\\
    A_{m1} &...& A_{1(j-1)} & A_{1(j+1)} & A_{mn}\\
    \end{bmatrix},
\end{equation}
where $\mathbf{A_{p1}}$ represents the  residue matrix after kicking off the $i^{th}$ line and the $j^{th}$ column, and all the residue matrices follow a similar format listed in Eq. (6).

The complexity of the algorithm is $O(\rm mn^{2})$, where it finds a solution about ordered paths of number $n$ in a capacity of $mn$ paths in a polynomial time, which belongs to the P problem. On the other hand, typical task assignment algorithms such as the Kuhn-Munkres algorithm perform complexity of $O(\rm n^{3})$. In the multi-UUV system, the number of the targets $m$ is always supposed to be smaller or equal to the number of vehicles $n$, meaning that $O(\rm mn^{2})$ is faster than $O(\rm n^{3})$ respectively. Therefore, the efficiency of the proposed task assignment method can be guaranteed.
%%%%%%%%%%%%%%%%%%%%%%
\subsubsection{Current Effect Elimination Component}
The current position of the vehicle decides its next position, represented by a single neuron in the neural network model that belongs to a single grid in the map. The resolution of the optimal path is deducted neuron by neuron, grid by grid, therefore, the effect of the ocean currents can be decomposed into grid units, as shown in Fig. 5. The different directions and magnitudes of the current velocity $\mathbf{v_{\rm cur}}$ (in blue) and the desired velocity of the vehicle $\mathbf{v_{\rm d}}$ (in black) forms the off-track actual velocity $\mathbf{v_{\rm act}}$ (in red) at each step, which drags the vehicle away from its supposed sailing trajectory and induces large deviations or unexpected collisions for the pathfinding task (see Fig. 5).

A third-party velocity vector, the adjustment velocity vector $\mathbf{v_{\rm adj}}$ (in green) is introduced to eliminate the off-track effect brought by the current velocity, which is produced by the vehicle based on the parallelogram law (Fig. 5). According to the parallelogram law, the summation result of $\mathbf{v_{\rm cur}}$ and $\mathbf{v_{\rm adj}}$ gives $\mathbf{v_{\rm d}}$, thus retaining the vehicle along the desired trajectory with both the desired direction and velocity. The magnitude of $\mathbf{v_{\rm d}}$ can be set initially and its direction is deducted by the intelligent BINN method, and $\mathbf{v_{\rm cur}}$ can be addressed based on the environment. Hence suppose the coordinates of $\mathbf{v_{\rm d}}$ is ($x_d$, $y_d$), and the $\mathbf{v_{\rm cur}}$ is ($x_{cur}$, $y_{cur}$), the position of the adjustment velocity vector can be obtained as
\begin{equation}
    \mathbf{v_{adj}}:(x_d-x_{cur}, y_d-y_{cur}),
\end{equation}
where $\mathbf{v_{\rm adj}}$ represents the position of the adjustment velocity vector, which completes the current effect elimination component in the CBNNTAP algorithm for the 2D condition by Eq. (7).

\begin{figure}[t]
\begin{center}
        \includegraphics[scale=0.7]{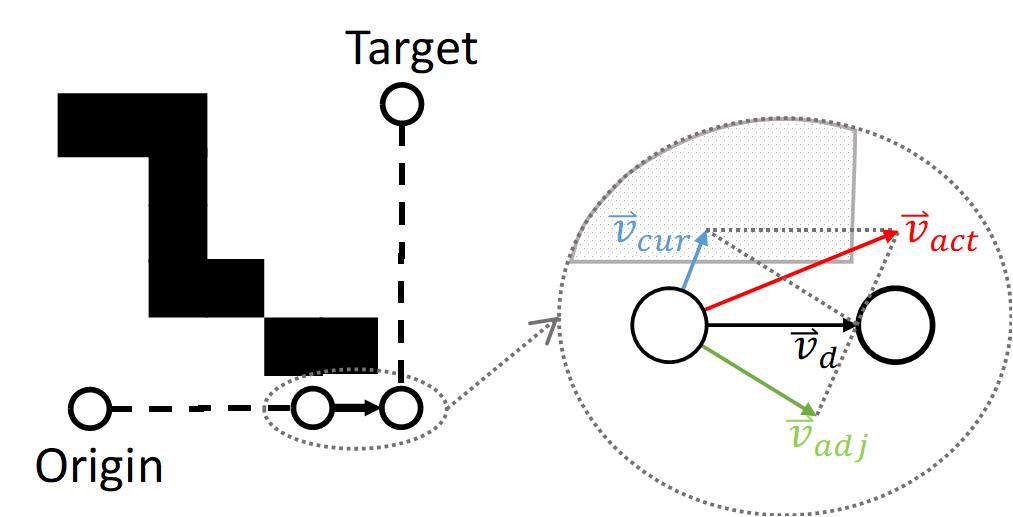}                         %\captionsetup{justification=centering}
        \caption*{Fig. 5. Elimination of the ocean current effect by introducing a third party vector $\mathbf{v_{\rm adj}}$ on a 2D map.}
\end{center} 
\end{figure}     %%%%%%true figure
A similar mechanism of elimination in the 3D model is applied, where one extra dimension is added to the system as well as the velocity vectors. The position of $\mathbf{v_{\rm adj}}$ can be attained based on the coordinates of $\mathbf{v_{\rm d}}$ and $\mathbf{v_{\rm cur}}$ based on their corresponding coordinates as
\begin{equation}
    \mathbf{v_{\rm adj}}: (x_d-x_{cur}, y_d-y_{cur}, z_d-z_{cur}),
\end{equation}
where $\mathbf{v_{\rm adj}}$ represents the position of the adjustment velocity vector, which forms the elimination component of the ocean current effect in the 3D condition by Eq. (8).

%%%%%%%%%%%%%%%%%%%%%%%%%%
\section{SIMULATION RESULTS AND ANALYSIS}
In this section, the target assignment and motion planning results of the CBNNTAP algorithm are given and compared with the algorithm that does not consider the effect of currents, such as the BNNP algorithm. Different conditions of current effect are applied, where the direction or velocity of the currents varies.
%%%%%%%%%%%%%%%%%%%%%%%%%%
\subsection{2D Simulation Results}
The target assignment and motion planning results of the multi-UUV system on the 2D map given by the CBNNTAP and the BNNP algorithms are shown in this section, with conditions of different current directions and velocities applied. The size of the map is $20 m \times 20 m$, desired velocity $v_{\rm d}$ of the vehicles are all set to be 1 m/s.

\subsubsection{Target Assignment}
In the 2D simulation, four UUVs waiting to be assigned, with their positions initially set at (9, 9), (6, 3), (2, 1) and (4, 17). The three targets are distributed at (5, 8), (19, 0) and (14, 14) (see Fig. 6). According to the algorithm explained in the previous section, the target assignment matrix can be derived based on the BNNP algorithm, where specific values of the matrix for the 2D simulation case are listed in the TABLE \uppercase\expandafter{\romannumeral1}. The values in the table will be denoted by "UUV-Target" to follow the matrix notation, for example, the value 0.1451 in TABLE \uppercase\expandafter{\romannumeral1} can be represented by "\uppercase\expandafter{\romannumeral1}-A".

\begin{table}[H]
\centering
\captionsetup{justification=centering}
\caption*{TABLE \uppercase\expandafter{\romannumeral1}. Normalized Distances of the Target Assignment in 2D Simulation (Assignment Matrix $\mathbf{A}$)}
\begin{tabular}{|c|c|c|c|c|}
\hline
\diagbox  {Target}{UUV} & \uppercase\expandafter{\romannumeral1} & \uppercase\expandafter{\romannumeral2} & \uppercase\expandafter{\romannumeral3} & \uppercase\expandafter{\romannumeral4} \\
\hline
A & 0.1451 & 0.1987 & 0.3261 & 0.3788\\
\hline
B & 0.5730 & 0.6335 & 0.8135 & 1\\
\hline
C & 0.2547 & 0.5994 & 0.7453 & 0.4611\\
\hline
\end{tabular}
\end{table}

The smallest normalized distance \uppercase\expandafter{\romannumeral1}-A, which refers to the path from UUV \uppercase\expandafter{\romannumeral1} to Target A is chosen at first due to its highest efficiency. Therefore, only UUV \uppercase\expandafter{\romannumeral2} to \uppercase\expandafter{\romannumeral4} are left for Target B and Target C, which means the first line and the first column of the matrix are no longer considered. Among the left choices, the most efficient path with the shortest normalized distances is \uppercase\expandafter{\romannumeral4}-C, meaning Target C is assigned to the UUV \uppercase\expandafter{\romannumeral4}, such that the last line and the last column of the matrix are excluded. Based on the same algorithm, the UUV \uppercase\expandafter{\romannumeral2} is assigned to Target B, as \uppercase\expandafter{\romannumeral2}-B ranks the first among the left matrix. The target assignment result is clearly shown in Fig. 6, where three chosen paths are noted by different marks.

\begin{figure}[t]
\begin{center}
    %\begin{flushmiddle}
        \includegraphics[scale=0.42]{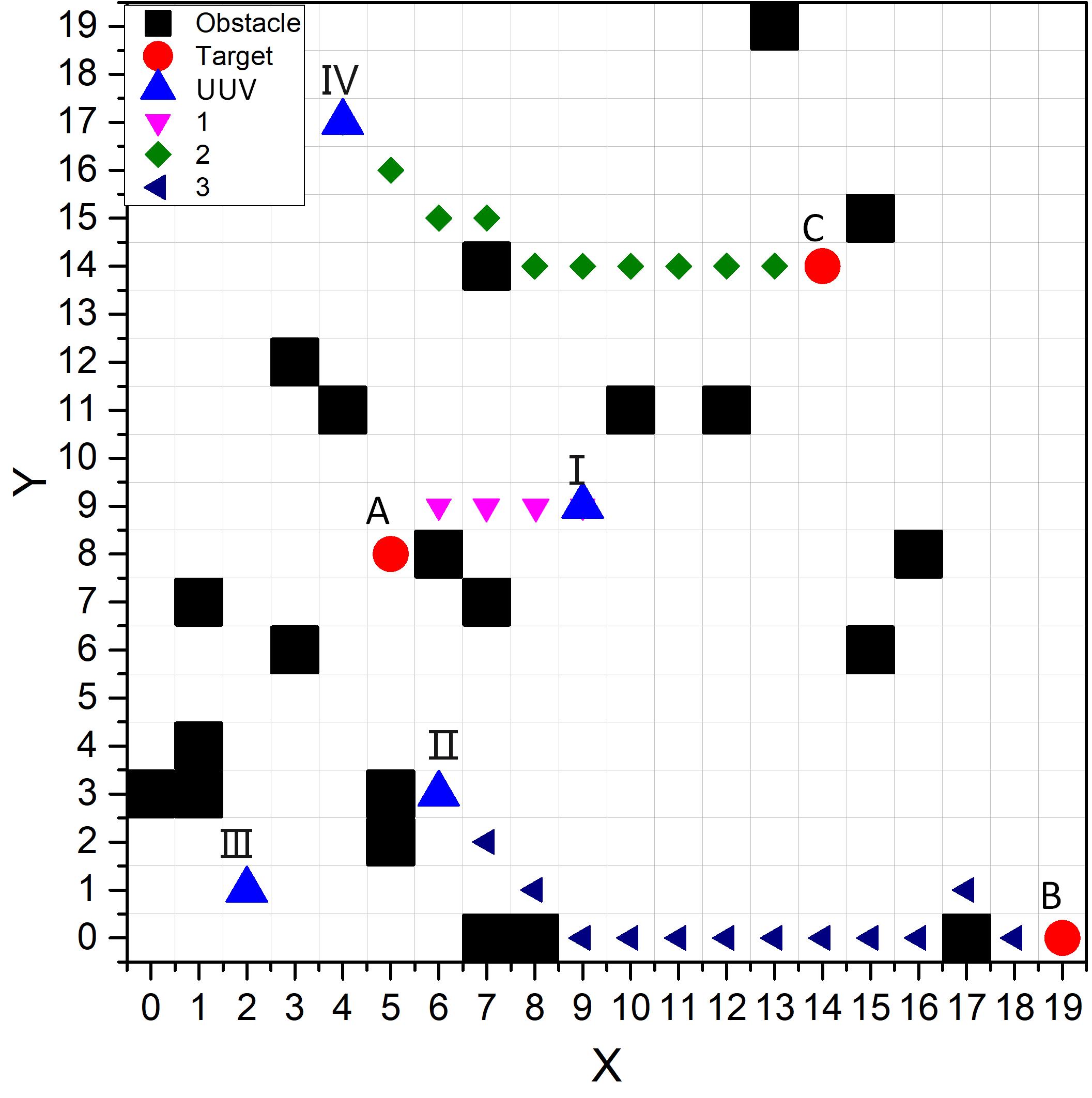}                         %\captionsetup{justification=centering}
        \caption*{Fig. 6. Target Assignment Result of the 2D simulation given by the CBNNTAP algorithm}
    %\end{flushmiddle}
\end{center} 
\end{figure}    %%%%%%true figure
%%%%%%%%%%%%%%%%%%%%%%
\subsubsection{Planning under Static Currents} 
In this section, the planning results of a multi-UUV system controlled by the proposed CBNNTAP and conventional BNNP method with the effect of static currents are presented and analyzed. 
\paragraph{Static Currents of Different Directions}
\begin{figure}[!b]
\begin{center}
    %\begin{flushmiddle}
        \includegraphics[scale=0.38]{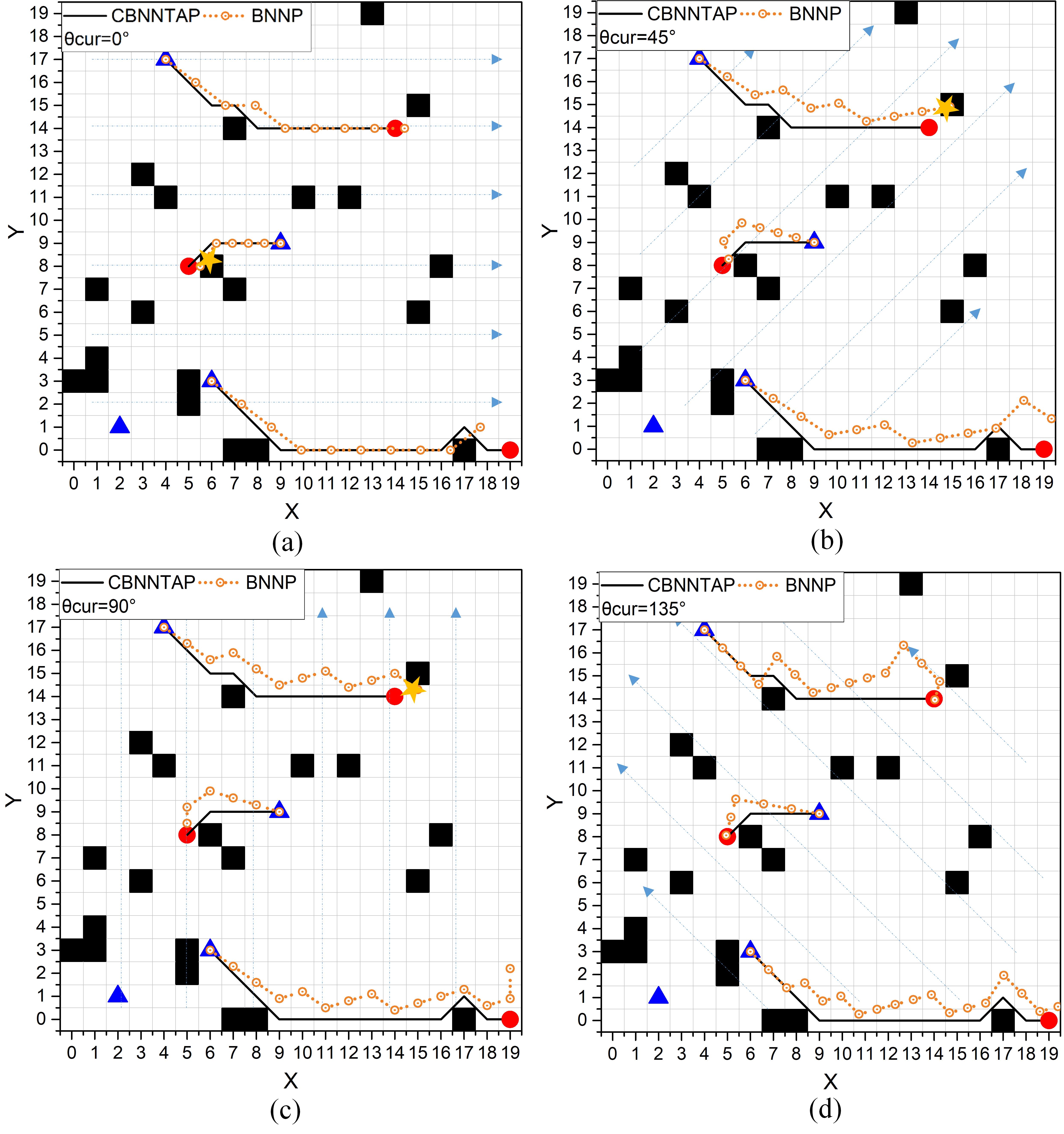}                         %\captionsetup{justification=centering}
        \caption*{Fig. 7. 2D comparison results of the paths given by the CBNNTAP algorithm and the BNNP algorithm under currents of different directions. The direction of the currents (all counterclockwise): (a) 0\degree; (b) 45\degree; (c) 90\degree; (d) 135\degree}	
    %\end{flushmiddle}
\end{center} 
\end{figure}    %%%%%%true figure

In this section, currents come in different directions of 0\degree, 45\degree, 90\degree and 135\degree counterclockwise are applied on the 2D map for evaluating the currents direction effect on the planning of the multi-UUV system, with the currents speed consistently set at 0.3m/s. Deviations are produced in the BNNP results (in orange) compared to the optimal planning path (in black line), see Fig. 7. Moreover, the change of deviations shown in the BNNP algorithm follows the change of the current direction. For instance, when the ocean current direction is farther from the vehicle's supposed heading direction to the target, the deviation of the BNNP algorithm appears to be larger as the currents offer less assistance to the vehicle for reaching the target, and vice versa. Even under cases that the current direction helps some of the vehicles heading to their destination (i.e. \uppercase\expandafter{\romannumeral4}-C and \uppercase\expandafter{\romannumeral2}-B in Fig. 7 (a))  while drags others away from the target (\uppercase\expandafter{\romannumeral1}-A in Fig. 7 (a)), unexpected collisions cannot be avoided by the BNNP algorithm (yellow star in \uppercase\expandafter{\romannumeral2}-B). As a comparison, the CBNNTAP algorithm successfully sustains the multiple vehicles along the optimal paths under all circumstances and avoids the possibility of bumping into underwater obstacles, whenever the direction of currents varies, which verifies the effectiveness of the proposed algorithm. 
%%%%%%%%%%%%%%%%%%%%%%%
\paragraph{Static Currents of Different Velocities}
\begin{figure}[t]
\begin{center}
    %\begin{flushmiddle}
        \includegraphics[scale=0.38]{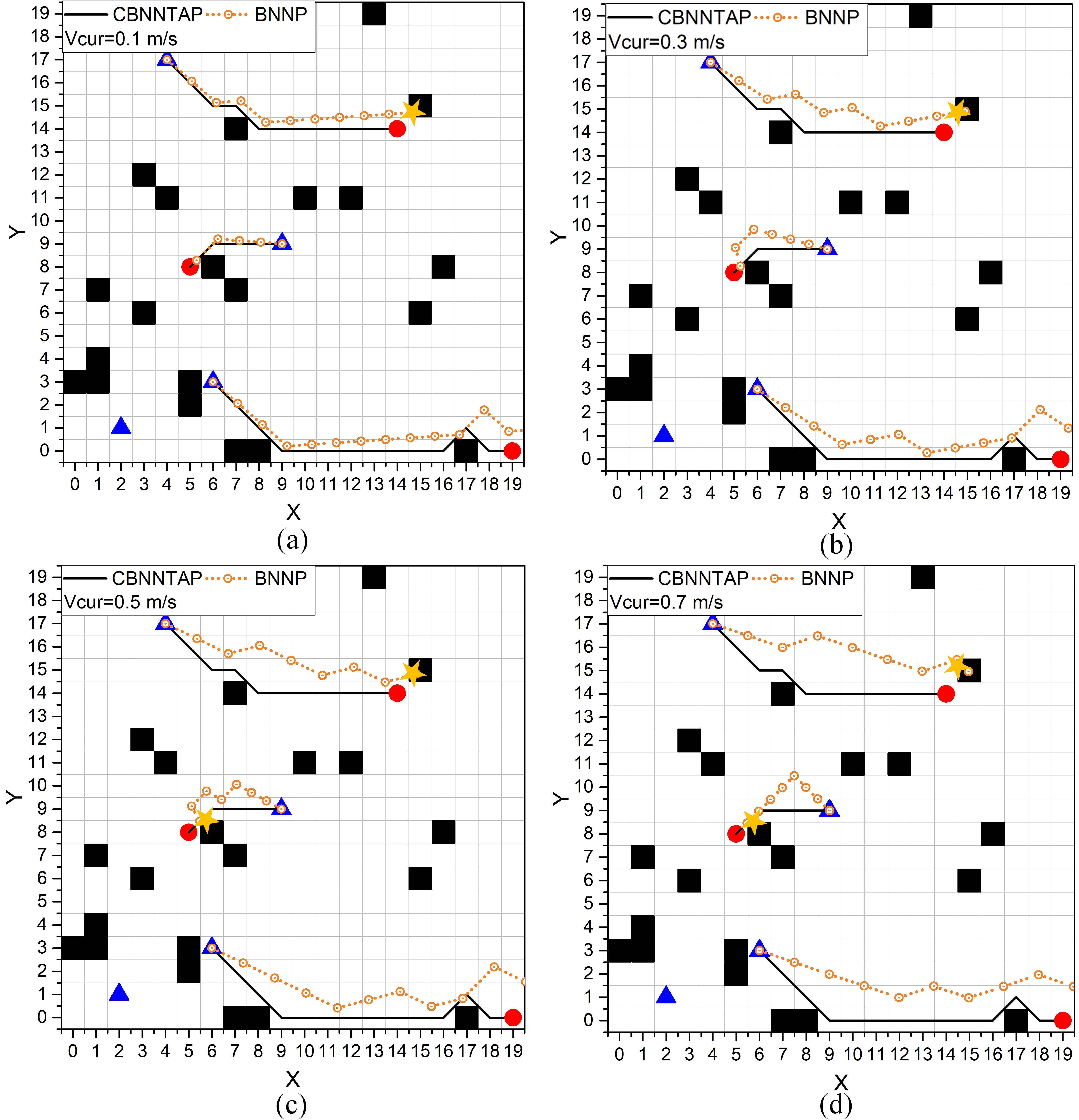}                         %\captionsetup{justification=centering}
        \caption*{Fig. 8. 2D comparison results of the paths given the CBNNTAP algorithm and the BNNP algorithm under currents of different velocities. Velocity of currents: (a) 0.1 m/s; (b) 0.3 m/s; (c) 0.5 m/s; (d) 0.7 m/s}	
    %\end{flushmiddle}
\end{center} 
\end{figure}    %%%%%%true figure

\begin{table*}[h]
    \centering
    \captionsetup{justification=centering}
    \caption*{TABLE \uppercase\expandafter{\romannumeral2}. 2D Planning Results under Effect of Static Currents\\(Unit: m. C: Collision, F: Fail to reach the destination without collision.)}
    \setlength{\tabcolsep}{0.2mm}{
    \begin{tabular}{|c|cc|cc|cc|cc|cc|cc|cc|}
    \hline
    &\multicolumn{8}{c|}{Direction variation ($\mathbf{v}_{\rm cur}=0.3 m/s$)} & \multicolumn{6}{c|}{Velocity variation ($\Theta_{\rm cur}=45\degree$)}\\
    \hline
    & \multicolumn{2}{c}{0°}& \multicolumn{2}{c}{45°} &  \multicolumn{2}{c}{90°} &  \multicolumn{2}{c|}{135°} &  \multicolumn{2}{c}{0.1 m/s} &  \multicolumn{2}{c}{0.5 m/s} & \multicolumn{2}{c|}{0.7 m/s}\\
    \hline
    \diagbox  {Path}{Algorithm} & CBNNTAP & BNNP  & CBNNTAP & BNNP & CBNNTAP & BNNP & CBNNTAP & BNNP & CBNNTAP & BNNP  & CBNNTAP & BNNP  & CBNNTAP & BNNP\\
    \hline
    I-A & 4.4142 & C & 4.4142 & 5.1944 & 4.4142 & 5.0527 & 4.4142 & 5.9524 & 4.4142 & 4.5105 & 4.4142 & C & 4.4142 & C\\
    \hline
    \uppercase\expandafter{\romannumeral4}-C & 10.2426 & 14.1204 & 10.2426 & C & 10.2426 & C & 10.2426 & 14.7721 & 10.2426 & C & 10.2426 & C & 10.2426 & C\\
    \hline
 \uppercase\expandafter{\romannumeral2}-B & 14.0711 & F  & 14.0711 & F & 14.0711 & F & 14.0711 & F & 14.0711 & F & 14.0711 & F & 14.0711 & F\\
    \hline
    \end{tabular}}
\end{table*}

Currents of different flow rates (represented by the current velocities) are applied in the 2D simulation of the multi-UUV motion planning, where its simulation results are presented in this section. 0.1 m/s, 0.3 m/s, 0.5 m/s and 0.7 m/s are separately assigned as the velocities of the currents, with the current direction consistently set at 45\degree counterclockwise. The BNNP paths (in orange) deviate with the flow speed, where the deviation enlarges when the speed increases  (see Fig. 8). While the CBNNTAP paths always keep track of the most optimal path as planned, even in the case of high flow rate (Fig. 8(d)). Hence the elimination of the effect of currents is successfully realized by the CBNNTAP algorithm, neglecting the change of the directions or velocities of the currents.

Based the path distances given in Table \uppercase\expandafter{\romannumeral2}, whenever the current direction or velocity varies, the path distances deducted by the CBNNTAP algorithm for all tasks always guarantees the shortest values, 4.4142 m for \uppercase\expandafter{\romannumeral1}-A, 10.2426 m for \uppercase\expandafter{\romannumeral4}-C and 14.0711 m for \uppercase\expandafter{\romannumeral2}-B, indicating the robust optimal motion planning results for the multi-UUV system. The BNNP paths generally consume around 30\%-50\% longer distances compared to the CBNNTAP paths. When the current velocity increases or the current direction pulls the vehicle away from the supposed direction, the vehicle of the BNNP method cannot accomplish the journey to the destination as desired by producing large deviations or collisions. The unpredictable failure of the BNNP algorithm points out the importance of current effect elimination in the multi-UUV system, which is successfully achieved by the CBNNTAP method.
%%%%%%%%%%%%%%%%
\begin{figure}[!b]
\begin{center}
        \includegraphics[width=0.45\textwidth]{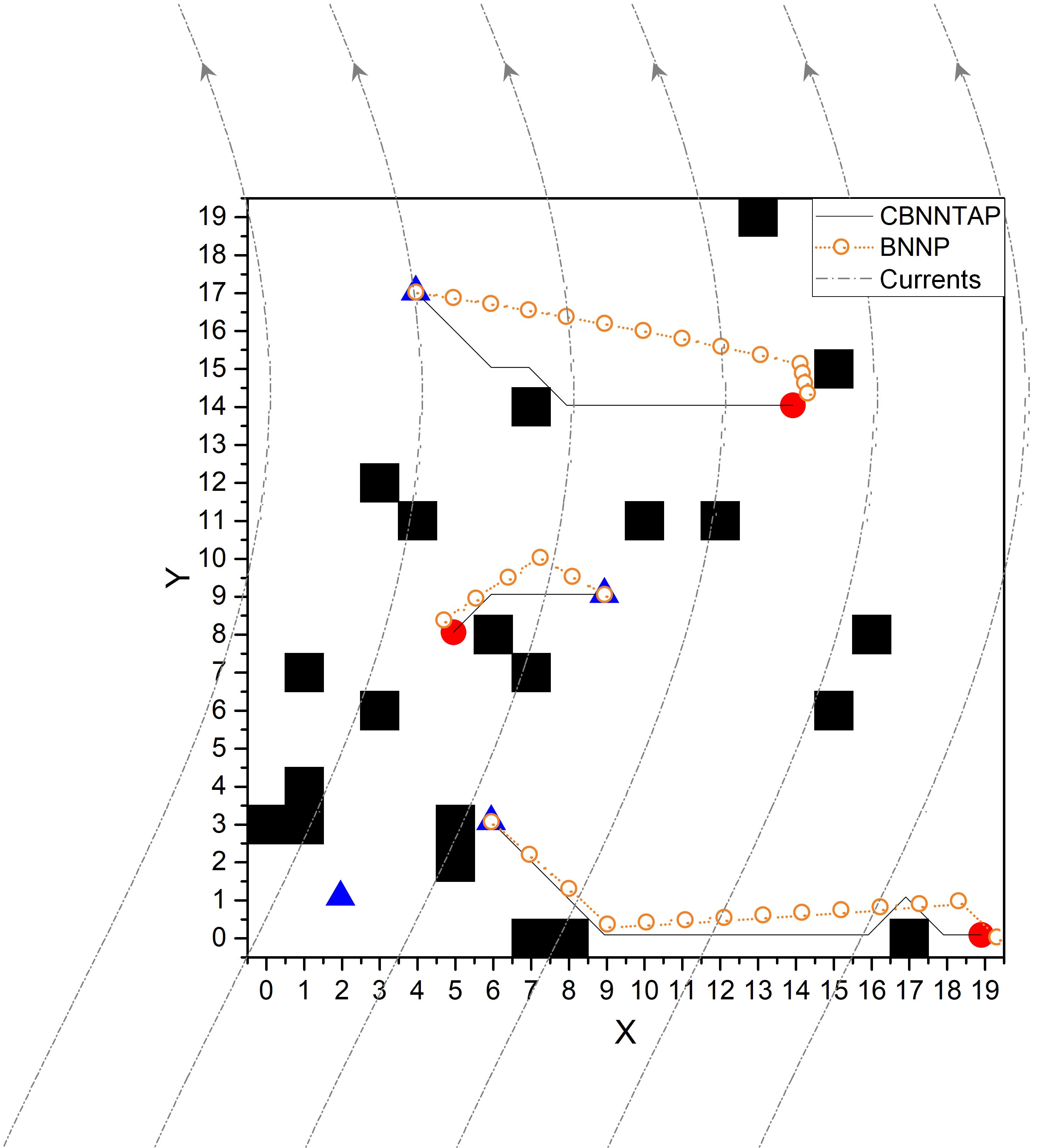}                         %\captionsetup{justification=centering}
        \caption*{Fig. 9. 2D planning results given by the CBNNTAP algorithm and BNNP algorithm with dynamic currents.}
\end{center} 
\end{figure}    %%%%%%true figure

\subsubsection{Planning under Dynamic Currents}
The target assignment and motion planning results under dynamic ocean currents with continuously changing velocities and directions are presented in Fig. 6. The dynamic currents come in the form of wave functions pointing from the bottom to the top as 
\begin{equation}
    x=5\sin(0.1y)+b,
\end{equation}
where $b$ represents a changing constant such that the dynamic currents can cover the whole map area. The direction of currents $\theta_{\vec{v}_{cur}}$ is along the tangent line of the wave function at each location.

The velocity of currents is set as
\begin{equation}
    v_{\vec{v}_{cur}}=(y+1)\times0.05,
\end{equation}
where the velocity gradually increases along the y-axis, and its unit in m/s.

The planning results of the BNNP algorithm vibrate fiercely with the current flowing tendency and finally consume a much longer distance when reaching the target as the current direction or velocity drags the vehicle far from its heading direction. Meanwhile, the CBNNTAP algorithm successfully retains optimal planning results for this multi-UUV system throughout the whole process, neglecting the dynamically changing currents. This supports the effectiveness of the CBNNTAP method for eliminating the dynamic current effect whenever the current direction or velocity is changing in a complex nonlinear form.

%%%%%%%%%%%%%%
\subsection{3D Simulation Results}
The 3D simulation results of the CBNNTAP and the BNNP methods are given and compared in this section. Velocity magnitude of $\mathbf{v_{\rm d}}$ is assigned as 1 m/s. The coefficient $k_f$ is 0.5. The size of the 3D map is $10 m\times10 m\times10 m$. 

\subsubsection{Target Assignment}
The task assignment of the 3D simulation is given in this subsection. Four UUVs waiting to be assigned, with their position initially set at (9, 7, 8), (6, 3, 3), (2, 1, 1) and (4, 1, 0). The three targets are distributed at (4, 4, 4), (5, 8, 1) and (9, 0, 0) (see Fig. 10). The specific values of the assignment matrix for the 3D simulation case are listed in the TABLE \uppercase\expandafter{\romannumeral3}. The values in the table are denoted in the same form as the previous section, for example, "\uppercase\expandafter{\romannumeral1}-A" represents 0.6982 in the table.

\begin{table}[H]
\centering
\captionsetup{justification=centering}
\caption*{TABLE \uppercase\expandafter{\romannumeral3}. Normalized Distances of the Target Assignment in 3D Simulation (Assignment Matrix $\mathbf{A}$)}
\begin{tabular}{|c|c|c|c|c|}
\hline
\diagbox  {Target}{UUV} & \uppercase\expandafter{\romannumeral1} & \uppercase\expandafter{\romannumeral2} & \uppercase\expandafter{\romannumeral3} & \uppercase\expandafter{\romannumeral4} \\
\hline
A & 0.6982 & 0.2507 & 0.4476 & 0.4810\\
\hline
B & 0.8234 & 0.5639 & 0.7562 & 0.7094\\
\hline
C & 1 & 0.4767 & 0.7854 & 0.5727\\
\hline
\end{tabular}
\end{table}

The smallest normalized distance \uppercase\expandafter{\romannumeral2}-A, which refers to the path from UUV \uppercase\expandafter{\romannumeral2} to Target A is chosen at first due to its highest efficiency. Therefore, the first line and the second column of the matrix are no longer considered. Among the left choices, the most efficient path with the shortest normalized distances is \uppercase\expandafter{\romannumeral4}-C, meaning Target C is assigned to the UUV \uppercase\expandafter{\romannumeral4}, such that the third line and the last column of the matrix are excluded. Based on the same algorithm, the UUV \uppercase\expandafter{\romannumeral3} is assigned to the Target B, as \uppercase\expandafter{\romannumeral3}-B ranks the first among the left matrix. The target assignment result is clearly shown in Fig. 10, where the three chosen paths are noted by different marks.

\begin{figure}[t]
\begin{center}
    %\begin{flushmiddle}
        \includegraphics[scale=0.48]{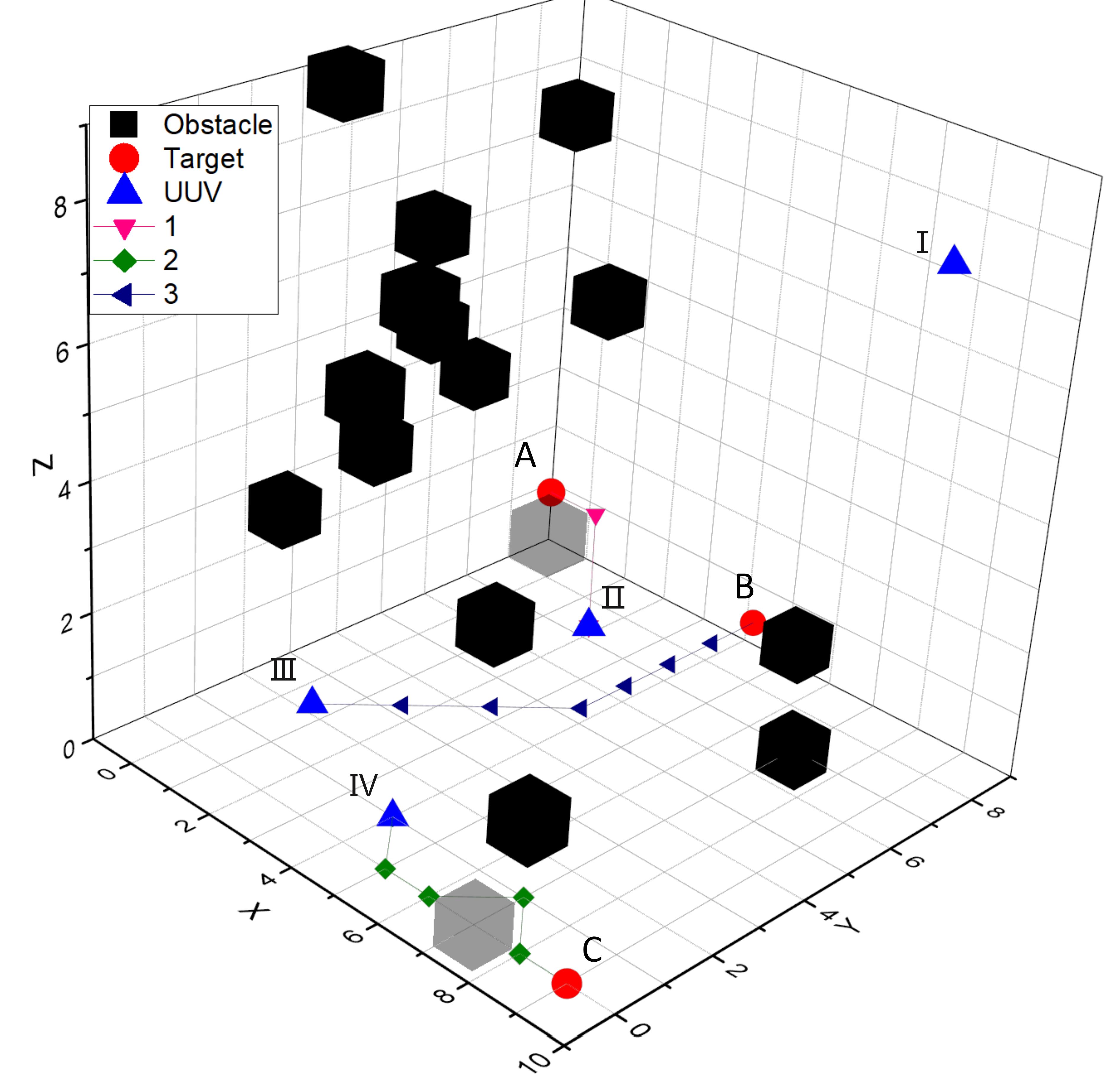}                         %\captionsetup{justification=centering}
        \caption*{Fig. 10. Target Assignment Result of the 3D simulation given by the CBNNTAP algorithm.}
    %\end{flushmiddle}
\end{center} 
\end{figure}    %%%%%%true figure
%%%%%%%%%%
\subsubsection{Planning under Dynamic Currents}

\begin{figure*}[t]
\begin{center}
    %\begin{flushmiddle}
        \includegraphics[scale=0.5]{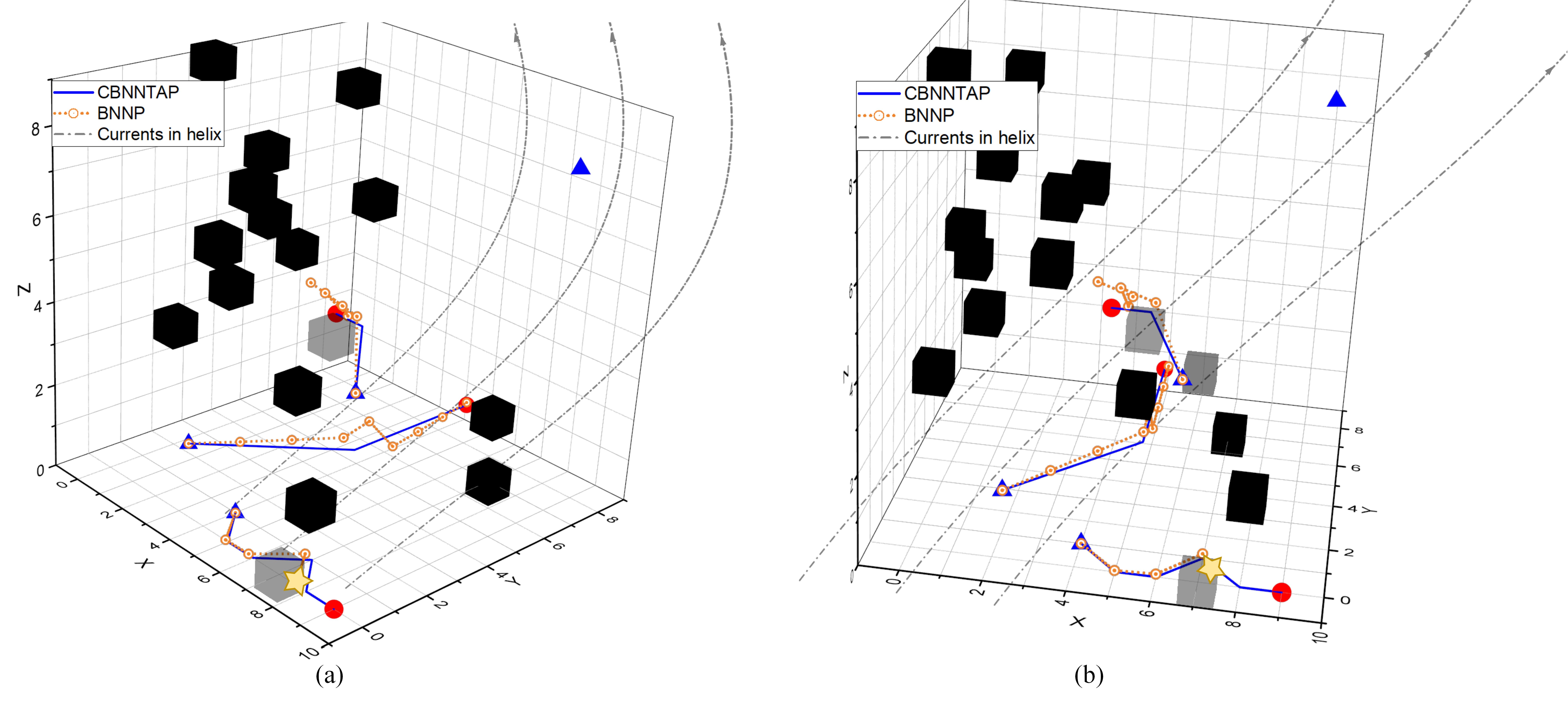}                         \captionsetup{justification=centering}
        \caption*{Fig. 11. 3D comparison results of the paths given by the CBNNTAP algorithm and the BNNP algorithm under dynamic currents.}	
    %\end{flushmiddle}
\end{center} 
\end{figure*}     %%%%%%true figure

The planning results in 3D condition of a multi-UUV system under ocean currents in a nonlinear form are presented in Fig. 11. The dynamic currents come in the form of 3D helices pointing from the bottom to the top as 
\begin{equation}
\left\{
\begin{array}{lr}
\begin{aligned}
    &x=10\sin(0.1t)+g,\\
    &y=10\cos(0.1t)+h,\\
    &z=t,
\end{aligned}
\end{array}
\right.\\
\end{equation}
where $t$, $g$ and $h$ represent changing constants such that the dynamic currents can cover the whole 3D map area. The direction of currents $\theta_{\vec{v}_{cur}}$ is along the tangent line of the helix at each location.

The velocity of currents is set as
\begin{equation}
    v_{\vec{v}_{cur}}=(z+1)\times0.1,
\end{equation}
where the velocity gradually increases along the z-axis.

The BNNP paths (in orange) show a fierce vibration with the trend of currents flowing, and even lose control before arriving at the supposed target (see \uppercase\expandafter{\romannumeral2}-A and \uppercase\expandafter{\romannumeral4}-C in Fig. 11). Moreover, a collision occurs in the BNNP results when an inevitable deviation is deduced by the dynamic ocean current effect (\uppercase\expandafter{\romannumeral4}-C in Fig. 11). The CBNNTAP algorithm retains the desired path (in black line) throughout the whole process for multiple vehicles without collisions, neglecting the changing flow rate. This indicates the success of the CBNNTAP algorithm in eliminating the current effect for the multi-UUV system in the 3D underwater environment of dynamic currents. 

\begin{table}[h]
    \centering
    \captionsetup{justification=centering}
    \caption*{TABLE \uppercase\expandafter{\romannumeral4}. 3D Planning Results under Effect of Dynamic Currents\\(Unit: m, C: Collision, F: Fail to reach the destination without collision.)}
    \setlength{\tabcolsep}{0.5mm}{
    \begin{tabular}{|c|c|c|}
    \hline
    \diagbox  {Path}{Algorithm} & CBNNTAP & BNNP  \\
    \hline
    \uppercase\expandafter{\romannumeral2}-A & 2.7321 & F \\
    \hline
    \uppercase\expandafter{\romannumeral4}-C & 6.2426 & C\&F \\
    \hline
 \uppercase\expandafter{\romannumeral3}-B & 8.2426 & 8.7356  \\
    \hline
    \end{tabular}}
\end{table}

%%%%%%%%%%%
Distances of the CBNNTAP and BNNP algorithms under the effect of dynamic currents are listed in Table \uppercase\expandafter{\romannumeral4}. The distance data derived by the CBNNTAP algorithm under the dynamic currents show that the vehicle completes the optimal target assignment and motion planning with the shortest distance in the 3D environment, regardless of the current direction or flow rate. The BNNP path cannot reach the destination in most cases and the deviation rises with the increment of the flow rate (\uppercase\expandafter{\romannumeral2}-A and \uppercase\expandafter{\romannumeral4}-C). Even in the BNNP cases where the vehicle completes the navigation (see \uppercase\expandafter{\romannumeral3}-B), the distance performs apparently longer than the path of 8.2426 m given by the CBNNTAP algorithm, with a longer journey of 8.7356 m, which supports the effectiveness of the proposed CBNNTAP algorithm in a multi-UUV system under 3D condition.

\section{Conclusion}
In this article, the problem of target assignment and motion planning under the effect of ocean currents for a multi-UUV system is discussed. A BINN-based algorithm (CBNNTAP) is proposed to eliminate the effect of currents in the system of multiple targets and vehicles. The adjusting component in the CBNNTAP algorithm deducts the corresponding velocity vector that the UUV should provide to balance off the deviation induced by the current effect and guarantee the vehicle navigates along the most optimal path. Therefore the target assignment can be addressed according to the pre-optimal path planning process, by conducting an assignment matrix that presents the efficiency of the task from each UUV to the separate target, whose priority order is decided based on the normalized path distances. The simulation results show that the proposed algorithm successfully realizes the current effect elimination, collision avoidance and the efficient collaboration of vehicles, whenever currents of different directions or velocities are applied. For future research, the challenges that are currently not involved such as handling uncertainty, managing inter-agent communication and coordination, and addressing energy consumption and potential conflicts will be considered.

% if have a single appendix:
%\appendix[Proof of the Zonklar Equations]
% or
%\appendix  % for no appendix heading
% do not use \section anymore after \appendix, only \section*
% is possibly needed

% use appendices with more than one appendix
% then use \section to start each appendix
% you must declare a \section before using any
% \subsection or using \label (\appendices by itself
% starts a section numbered zero.)
%

%\appendices
%\section{Proof of the First Zonklar Equation}
%Appendix one text goes here.

% you can choose not to have a title for an appendix
% if you want by leaving the argument blank
%\section{}
%Appendix two text goes here.

% use section* for acknowledgment
%%%%%%%%%%%%%%%%%%%%%%%%%%%%%%%%%%%%%%%%%%%%%%%%%%%%%%%%%%%%%%%%%%%%%%%%%%%%%%%%%%%%%%%%%%%%%%%%%%%%%%%%%%%%%%%%%
%\section*{Acknowledgment}
%The authors would like to thank...

% Can use something like this to put references on a page
% by themselves when using endfloat and the captionsoff option.
\ifCLASSOPTIONcaptionsoff
  \newpage
\fi

% trigger a \newpage just before the given reference
% number - used to balance the columns on the last page
% adjust value as needed - may need to be readjusted if
% the document is modified later
%\IEEEtriggeratref{8}
% The "triggered" command can be changed if desired:
%\IEEEtriggercmd{\enlargethispage{-5in}}

% references section

% can use a bibliography generated by BibTeX as a .bbl file
% BibTeX documentation can be easily obtained at:
% http://mirror.ctan.org/biblio/bibtex/contrib/doc/
% The IEEEtran BibTeX style support page is at:
% http://www.michaelshell.org/tex/ieeetran/bibtex/
%\bibliographystyle{IEEEtran}
% argument is your BibTeX string definitions and bibliography database(s)
%\bibliography{IEEEabrv,../bib/paper}
%
% <OR> manually copy in the resultant .bbl file
% set second argument of \begin to the number of references
% (used to reserve space for the reference number labels box)

%\printbibliography
\scriptsize
\bibliographystyle{IEEEtran}
%\addbibresource{IEEEabrv.bib}

%%%%references 11.28
\bibliography{ref}    %%%%%true ref

\end{document}